\documentclass[letterpaper, 10 pt, conference]{ieeeconf}

\IEEEoverridecommandlockouts
\usepackage{times}
\usepackage{epsfig}
\usepackage{graphicx}
\usepackage{amsmath}
\usepackage{amssymb}
\usepackage{multicol}
\usepackage{rotating}
\usepackage{multirow}
\usepackage{flushend}
\usepackage{float}
\usepackage{amsmath,amssymb,amsfonts}
\usepackage{siunitx}                    
\usepackage{dsfont,latexsym,cite, comment, graphicx,color}
\usepackage[utf8]{inputenc}  
\usepackage{url}
\usepackage{graphicx}
\usepackage{graphics}
\usepackage[dvipsnames]{xcolor}
\definecolor{car}{RGB}{100, 150, 245} 
\definecolor{bicycle}{RGB}{100, 230, 245} 
\definecolor{motorcycle}{RGB}{30, 60, 150} 
\definecolor{truck}{RGB}{80, 30, 180} 
\definecolor{other-vehicle}{RGB}{75, 0, 175} 
\definecolor{person}{RGB}{0, 0, 255} 
\definecolor{bicyclist}{RGB}{0, 0, 150} 
\definecolor{motorcyclist}{RGB}{0, 0, 100} 
\definecolor{road}{RGB}{255, 0, 255} 
\definecolor{parking}{RGB}{255, 150, 255} 
\definecolor{sidewalk}{RGB}{75, 0, 75} 
\definecolor{other-ground}{RGB}{175, 0, 75} 
\definecolor{building}{RGB}{255, 200, 0} 
\definecolor{fence}{RGB}{255, 120, 50} 
\definecolor{vegetation}{RGB}{0, 175, 0} 
\definecolor{trunk}{RGB}{255, 150, 0} 
\definecolor{terrain}{RGB}{150, 240, 80} 
\definecolor{pole}{RGB}{255, 240, 150} 
\definecolor{traffic-sign}{RGB}{255, 0, 0} 
\definecolor{wire}{RGB}{255, 0, 0} 
\definecolor{ground}{RGB}{150, 240, 80} 
\usepackage{pifont}

\usepackage{subfigure}
\usepackage{algorithm}
\usepackage[noend]{algpseudocode}

\newcommand{\compressParag}{\looseness = -1} 

\usepackage[acronym]{glossaries}
\glsdisablehyper
\setacronymstyle{long-short}
\newacronym{ss}{SS}{Semantic Segmentation}
\newacronym{som}{SOM}{Semantic Occupancy Mapping}
\newacronym{ssc}{SSC}{Semantic Scene Completion}
\newacronym{nerf}{NeRF}{Neural Radiance Field}
\newacronym{uav}{UAV}{Unmanned Aerial Vehicle}
\newacronym{slam}{SLAM}{Simultaneous Localization and Mapping}
\newacronym{da}{DA}{Domain Adaptation}
\newacronym{tta}{TTA}{Test-Time Adaptation}
\newacronym{vfm}{VFM}{Vision Foundation Model}
\newacronym{rv}{RV}{Range-View}
\newacronym{bev}{BEV}{Birds-Eye-View}
\newacronym{tpv}{TPV}{Tri-Perspective-View}
\newacronym{miou}{$mIoU$}{Mean Intersection Over Union}
\newacronym{cv}{CV}{Computer Vision}
\newacronym{nlp}{NLP}{Natural Language Processing}
\newacronym{llm}{LLM}{Large Language Model}
\newacronym{cmst}{3D-CN}{3D Consistency Network}

\usepackage{arydshln}
\makeatletter
\let\NAT@parse\undefined
\makeatother
\usepackage[colorlinks,citecolor=red,urlcolor=blue,bookmarks=false,hypertexnames=true]{hyperref}
\title{\LARGE \bf LeAP: Consistent multi-domain 3D labeling using Foundation Models}

\author{Simon Gebraad$^{1}$\\
\and
Andras Palffy$^{1,2}$\\
\and
Holger Caesar$^{1}$%
\thanks{$^{1}$ Authors are or where with Delft University of Technology. {\tt j.s.gebraad@student.tudelft.nl}}%
\thanks{$^{2}$ Authors are with PercivAI.}%
}
\begin{document}
\maketitle

\begin{abstract}
Availability of datasets is a strong driver for research on 3D semantic understanding, and whilst obtaining unlabeled 3D point cloud data is straightforward, manually annotating this data with semantic labels is time-consuming and costly. 
Recently, \glspl{vfm} enable open-set semantic segmentation on camera images, potentially aiding automatic labeling. However, \glspl{vfm} for 3D data have been limited to adaptations of 2D models, which can introduce inconsistencies to 3D labels. This work introduces \textbf{Label Any Pointcloud (LeAP)}, leveraging 2D \glspl{vfm} to automatically label 3D data with \textit{any} set of classes in \textit{any} kind of application whilst ensuring label consistency. Using a Bayesian update, point labels are combined into voxels to improve spatio-temporal consistency. A novel \gls{cmst} exploits 3D information to further improve label quality.
Through various experiments, we show that our method can generate high-quality 3D semantic labels across diverse fields without \textit{any} manual labeling. Further, models adapted to new domains using our labels show up to a $34.2$ mIoU increase in semantic segmentation tasks.

\end{abstract}

\section{Introduction}
In recent years, machine perception has developed rapidly, supported by advances in deep learning that have led to various models for 3D perception tasks. Labeled data is crucial for the development of these deep learning models. However, manually labeling 3D data with the required semantic labels is time-consuming and thereby expensive. Consequently, these models have mainly been developed for the well-funded urban automotive domain, where multiple extensive labeled datasets with synchronized multi-modal sensors are available, such as nuScenes~\cite{caesar_nuscenes_2020}, Waymo~\cite{sun_scalability_2020}, KITTI~\cite{geiger_are_2012}, SemanticKITTI~\cite{behley_semantickitti_2019} and KITTI-360~\cite{liao_kitti-360_2023}. 

\begin{figure}[h]
    \raggedright
    \includegraphics[width=0.47\textwidth]{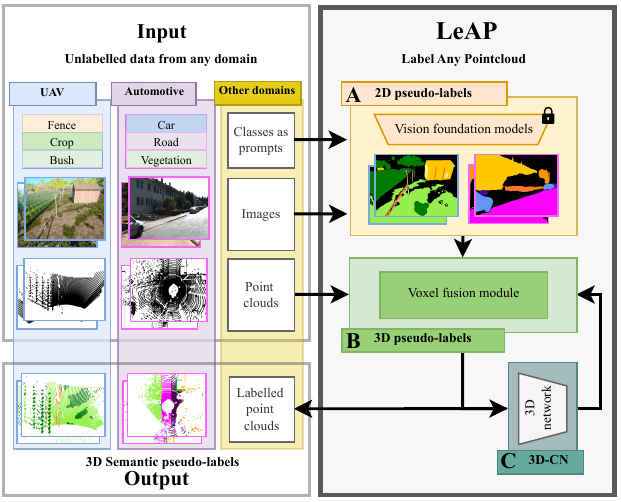}
    \caption{Overview of our LeAP automatic labeling method. ($\textbf{A}$) Taking only paired image-LiDAR data as input, foundation models are used to generate image labels for \textit{any} set of classes in \textit{any} application. ($\textbf{B}$) A Bayesian voxel update and ($\textbf{C}$) a novel \glsreset{cmst}\gls{cmst} improve label consistency, resulting in high quality pseudo-labels.}
    \label{overview}
\end{figure}

Recently, \textit{foundation models} have been introduced, which are large-scale neural network architectures trained on vast amounts of diverse data. This allows them to capture rich semantic representations of language or visual information, enabling strong generalization. Hence, these models can serve as foundational building blocks for downstream tasks~\cite{zhu_ponderv2_2023}, such as automatic labeling \cite{liu_offline_2024, karnchanachari_towards_2024}. They have seen extensive development in 2D \gls{cv}, resulting in \glspl{vfm} such as CLIP~\cite{radford_learning_2021}, SAM~\cite{kirillov_segment_2023} and Depth Anything~\cite{yang_depth_2024}. 
However, despite attempts to transfer 2D \glspl{vfm} to 3D~\cite{zhang_pointclip_2022, peng_openscene_2023, chen_clip2scene_2023, tan_ovo_2023, vobecky_pop-3d_2024, hess_lidarclip_2024}, \glspl{vfm} trained natively on 3D data are largely absent due to the limited scale and diversity of labeled 3D datasets~\cite{zhu_ponderv2_2023, wu_towards_2023}.  Errors in 2D-3D projection combined with the inherent lack of geometric awareness of 2D \glspl{vfm} can introduce inconsistencies into 3D adaptations. Exacerbating inconsistency, the output of \glspl{vfm} varies significantly depending on the prompt and visual context, which limits their effectiveness for automatic 3D labeling.

In this work, we address the consistency challenges associated with using 2D \glspl{vfm} for 3D labeling while preserving their open-set capabilities. Using unlabeled image-pointcloud pairs, we can generate high-quality 3D semantic pseudo-labels (i.e., machine-generated) for any set of classes (see Fig.~\ref{overview}). We use Bayesian updating to combine class labels over time into voxels, ensuring spatio-temporal consistency (Fig.~\ref{overview}\textbf{B}). Voxels also enable fusion with our novel \gls{cmst}, which further improves the geometric consistency (Fig.~\ref{overview}\textbf{C}) of our labels.
We make the following contributions:
\begin{itemize}
\item We introduce LeAP, a novel \textbf{domain agnostic} semantic pseudo-labeling tool for 3D data that leverages 2D open-vocabulary foundation models to work for \textit{any} arbitrary list of classes, in \textit{any} domain. 
\item In contrast to previous automatic labeling methods using \glspl{vfm}, we aggregate labels in voxels using the statistically grounded Bayesian update, which, in combination with our novel \glsreset{cmst} \gls{cmst} significantly improves \textbf{geometric and temporal consistency}.
\item We extensively evaluate the \textbf{multi-domain} capabilities of our method on an existing automotive dataset, and on our novel synthetic dataset for the less explored \gls{uav} domain.
\end{itemize}

\section{Related work}
{\begin{figure*}
\includegraphics[width=\textwidth]{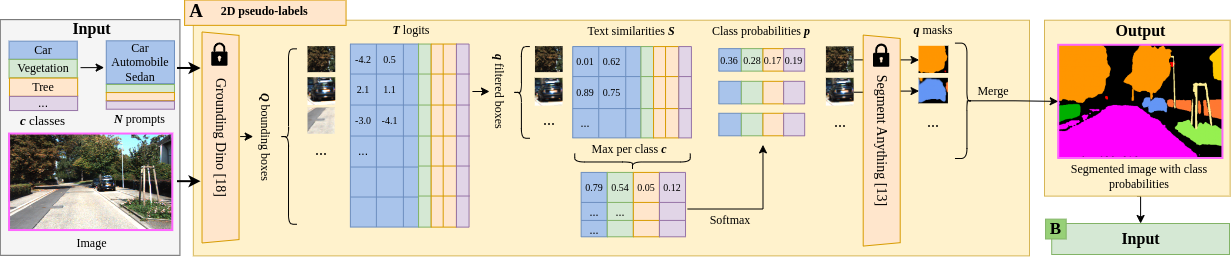}
\caption{The process of generating 2D pseudo-labels. Using unlabeled images and a list of classes, we use Grounding Dino~\cite{liu_grounding_2023} features to obtain regions with soft labels. Segment Anything~\cite{kirillov_segment_2023} converts these to detailed masks and allows us to obtain per-pixel soft labels.}
\vspace*{-\baselineskip}
\label{2dlabels}
\end{figure*}}

Supervised methods (e.g. ~\cite{maturana_voxnet_2015, graham_3d_2018, vedaldi_searching_2020, zhu_cylindrical_2021, zhao_svaseg_2022, lai_spherical_2023, zhang_occformer_2023, cao_monoscene_2022, wu_squeezesegv2_2018, milioto_rangenet_2019, cortinhal_salsanext_2020, zhang_polarnet_2020, kong_rethinking_2023, roldao_lmscnet_2020, cheng_s3cnet_2020, zuo_pointocc_2023, huang_tri-perspective_2023, qi_pointnet_2017, charles_pointnet_2017, thomas_kpconv_2019, hu_randla-net_2020, wu_point_2022, cheng_af_2021, xu_rpvnet_2021, ye_lidarmultinet_2023, lu_lidar-camera_2024}) for the online prediction of 3D semantics require large amounts of labeled 3D data which is often not available for novel application domains. 
Some approaches ~\cite{zhang_growsp_2023, liu_u3ds3_nodate} attempt unsupervised 3D \gls{ss}. However, the lack of semantic information in 3D features limits performance. Others~\cite{xie_pointcontrast_2020, zhang_self-supervised_2021, nunes_segcontrast_2022, zhu_ponderv2_2023} instead focus on unsupervised representation learning by pre-training on unlabeled LiDAR data, but these methods still require labels for effective fine-tuning. 
Previous research has used the extensive body of work in 2D \gls{cv} to overcome these issues.

\subsection{2D supervised 3D semantic understanding}
Various approaches~\cite{hayler_s4c_2024, zhang_occnerf_2023, genova_learning_2021, bultmann_real-time_2023, sautier_image--lidar_2022, mahmoud_self-supervised_2023, liu_segment_2023, vora_pointpainting_2020} use the well-researched 2D domain to enhance performance on 3D semantic tasks without relying on 3D labels.
Some works~\cite{genova_learning_2021, bultmann_real-time_2023, hayler_s4c_2024, reichardt_360_2023} use off-the-shelf pre-trained 2D semantic segmentation networks to supervise the training of 3D networks (so-called 'shelf-supervised'). PointPainting~\cite{vora_pointpainting_2020} uses 2D-3D projection to apply labels obtained from images to 3D points, however, projected labels are limited to the camera frame and can be noisy due to small errors in projection and masking. \cite{genova_learning_2021, reichardt_360_2023} nevertheless show that it is possible to effectively train 3D models using noisy projected labels as pseudo-labels by label filtering. Alternatively, ~\cite{hayler_s4c_2024, zhang_occnerf_2023} use \glspl{nerf} instead of projection to bridge the gap from 2D to 3D.  They utilize pre-trained 2D semantic segmentation networks and temporal consistency for semantic and depth supervision to train unsupervised \gls{ssc} models. 
However, these methods are constrained by their dependence on pre-trained, closed-set 2D models, which are often not available for novel domains.
Other studies~\cite{sautier_image--lidar_2022, mahmoud_self-supervised_2023} focus on representation learning with unlabeled camera-LiDAR data. The camera is used to group visually similar regions into superpixels. This knowledge is transferred to 3D, improving representations for 3D semantic tasks, though labels are still required for fine-tuning. 

\subsection{Foundation models for 3D semantics}
In recent years, various \glspl{vfm} have been introduced. Models like CLIP~\cite{radford_learning_2021} and Grounding Dino~\cite{liu_grounding_2023} combine language and vision for open-vocabulary image labeling and object detection respectively. However, these do not provide detailed pixel-wise labels and generally output only a single class per image or region. Other models like SAM~\cite{kirillov_segment_2023} (image segmentation) and Depth Anything~\cite{yang_depth_2024} (depth estimation) give per-pixel labels but lack semantics. Additionally, foundation models for 3D data are largely absent. Still, various methods exploit the open-set capabilities of 2D \glspl{vfm} for 3D semantic tasks.
Recent work~\cite{liu_segment_2023} based on~\cite{sautier_image--lidar_2022, mahmoud_self-supervised_2023} has employed segmentation \glspl{vfm} like SAM~\cite{kirillov_segment_2023} to improve representation learning by generating more consistent superpixels. However, the lack of semantic labels in the superpixels means labeled data is required for fine-tuning.
Some works~\cite{zhang_pointclip_2022, peng_openscene_2023, chen_clip2scene_2023, tan_ovo_2023, vobecky_pop-3d_2024, hess_lidarclip_2024} use CLIP~\cite{radford_learning_2021} to distill language features into 3D segmentation networks, enabling open-vocabulary capabilities in 3D applications. 
Although this allows these models to be highly flexible and predict any semantic class at test time, their universality also limits performance. Trained for image captioning, CLIP’s general language features are less suited for precise segmentation. Aggregating these features in 3D is also non-trivial which reduces temporal and geometric consistency. This limits the usefulness of these models for providing high quality labels. In contrast, we use \glspl{vfm} specialized for segmentation, using the statistically grounded Bayesian update in combination with our \gls{cmst} to improve label consistency.
Other works~\cite{khurana_shelf-supervised_2024, zhang_sam3d_2024, najibi_unsupervised_2023} use foundation models for pseudo-labeling to improve 3D object detection. However, as opposed to \gls{ss}, object detection does not require per-point labels, as it only uses course bounding boxes.
Most comparable to our work,~\cite{zhou_openannotate3d_2024, zhou_openannotate2_2024} also use open-vocabulary models for 3D semantic pseudo-labeling. However, the focus of their work is to simplify the workflow of human annotators by using \glspl{llm} to enable frame-by-frame annotation based on voice or text-prompts, requiring a human in-the-loop for supervision and label corrections. Hence, automatic temporal consistency is not considered, and evaluations on label quality across various domains are limited. We instead use a Bayesian voxel update to combine semantic labels and ensure temporal consistency, and evaluate label quality quantitatively across diverse domains.

\section{Method}

In this section we describe our approach to generate high quality 3D point-wise labels for any desired set of classes in any domain using only unlabeled camera-LiDAR data. We first cover how we use foundation models to generate soft 2D labels (Fig.~\ref{overview}$\textbf{A}$), and then how we use voxels for spatial-temporal accumulation to produce high quality 3D pseudo-labels (Fig.~\ref{overview}$\textbf{B}$). Finally, we highlight how our voxel-based approach enables modular integration of multiple sources of semantic labels by fusing the output of a self-trained 3D backbone with our camera-based pseudo-labels, which can further enhance pseudo-label quality (Fig.~\ref{overview}$\textbf{C}$).

\subsection{2D pseudo-label generation}
\label{subsec:2d}
To improve label consistency and enable Bayesian updating in our 3D labeling, we require per-pixel soft labels (i.e. probabilities). Hence, we assemble and modify the outputs of multiple foundation models to obtain detailed pixel-wise soft labels, illustrated in Fig.~\ref{2dlabels}. 

We first input an unlabeled image and a prompt into the pre-trained Grounding Dino~\cite{liu_grounding_2023} \gls{vfm} to obtain labels for bounding box regions. Specifically, given $c$ desired classes, we manually expand the prompt using three complementary strategies, namely (1) synonymous substitution, e.g., extending \textit{car} with \textit{automobile}, (2) adding additional categories according to the class descriptions to aid differentiation, e.g., adding \textit{van} to \textit{car}, and (3) replacing ambiguous classes with more detailed descriptions, e.g., replacing the \textit{other-vehicle} class with \textit{bus, train}, etc. This results in $n_c$ prompts for each class, with a total of $N = \sum^c_{i=1} n^i_c$ prompts. The pre-trained \gls{vfm} embeds these prompts into $T$ text tokens and outputs a logit vector  $\textbf{L}$ of size $Q \times T$, where $Q$ is a hyperparameter representing the number of query regions (i.e. bounding box proposals) of the image. The sigmoid of each logit represents the similarity of a region with a text embedding, which we use as a proxy for confidence. We filter regions where the maximum similarity is below a threshold hyperparameter, yielding a filtered similarity vector $\textbf{S}$ of size $q \times T$. For remaining regions, we extract the maximum similarity for each class from all prompts as $s(q,c) = \max_{T \in {1, 2, ..., n_c}} \textbf{S}(q)$. The softmax is applied to the original logits $\textbf{L}$ of these $c$ values to obtain class probabilities for each region $\textbf{p}(q,c)$.

To obtain detailed masks, we use the $q$ bounding boxes with class probabilities $\textbf{p}(q,c)$ as input for SAM~\cite{kirillov_segment_2023}. For regions with overlapping masks, we compute a weighted average of class probabilities, giving more weight to masks with higher similarity scores (confidence). This process results in a class probability distribution for each pixel $\textbf{p}(u,v)$ in the input camera image, where $u,v$ are image coordinates.

\subsection{3D pseudo-label generation}
\label{subsec:3d}

We subsequently use the soft 2D labels to obtain point-wise 3D semantic labels, see Fig.~\ref{3dlabels}$\textbf{B}$.
Using the known intrinsic camera properties and extrinsic transformation from LiDAR to camera, 3D points $\textbf{P}$ are projected onto the image plane, obtaining their image coordinates $(\textbf{u}, \textbf{v})$. Adapting the approach from PointPainting~\cite{vora_pointpainting_2020}, each point is then augmented with the class probability distribution to obtain $\textbf{P} = \begin{bmatrix}\textbf{x} & \textbf{y} & \textbf{z} & \textbf{p}(\textbf{u}, \textbf{v}) \end{bmatrix}^T$.

Errors in calibration and masking can cause points to be assigned incorrect labels. For example, labels of foreground objects are often assigned to points behind the object. Using the intuition that points within a 2D mask should also be close in 3D space, for each mask we cluster the points based on their distance from the camera. We then filter points that are not a part of the largest cluster.  Although this reduces the number of labeled points, we find it improves label quality.

To consistently combine the potentially ambiguous and noisy projected labels over time, we are inspired by work on \gls{slam} \cite{mccormac_semanticfusion_2017} and make use of the Bayesian update.  Rather than refining on the point-level, we make use of voxels. These serve as a dense, universal representation of 3D space, which enables the fusion of multiple point labels into a single voxel. Voxels also allow us to efficiently keep a memory of all past labels which enables retro-active labeling of points outside the camera fustrum.
To handle the potentially very large extent of the mapped 3D space, and thus the required memory, we use sparse voxel hashing \cite{niesner_real-time_2013}, allowing for efficient scaling. Following~\cite{bultmann_real-time_2023}, each voxel probabilistically fuses all point-wise soft labels within it using Bayes' Rule to obtain a statistically grounded probability distribution $\textbf{V}$ for each voxel.
Given an observed point $X_{k}$ with probability $p_i$ for class $i$, and a voxel $n$ with probabilities based on previous observations $V_n(p_i | X_{1:k-1})$, each voxel is updated using Eq.~\ref{eq:bayes}. This update scheme enables us to efficiently combine labels over time without explicitly keeping a memory of all points, whilst dealing with the ambiguity and noise from the projected labels.
\begin{equation}
    \label{eq:bayes}
    V_n(p_i | X_{1:k}) = \frac{V_n(p_i | X_{1:k-1})P(p_i | X_{k})} {\sum_{i}{V_n(p_i | X_{1:k-1})P(p_i | X_{k})}}
\end{equation}

Using Eq.~\ref{eq:bayes}, we combine all observations over time into a single voxel grid.
To enhance spatial consistency, we further refine the final grid using distance-weighted $k$-nearest averaging~\cite{reichardt_360_2023}, smoothing the class probabilities of each voxel as:
\begin{equation}
    \bar{p}_i = \sum_{m \in \mathcal{N}_k(n)} w_{nm}\; p_{im}
\end{equation}
Here, $\mathcal{N}_k(n)$ is the set of $k$-nearest neighbors of voxel $n$ and $\textbf{w}_n=\text{softmax}(-\textbf{d}_n)$ with  $\textbf{d}_n$ the distance vector of the $k$-nearest neighbors. 
Finally, to output per-point labels, we determine for each point in a point cloud the corresponding voxel label.

\subsection{Improving labels through a 3D consistency network}
\label{sec:self-training}
\begin{figure}[t]
    \centering
    \includegraphics[width=0.45\textwidth]{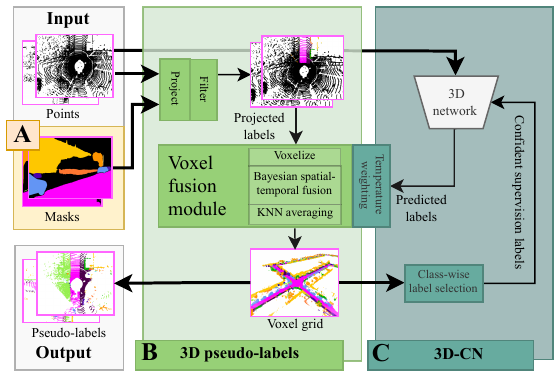}
    \caption{The process of generating 3D pseudo-labels. Point clouds are painted with image-based labels and probabilistically accumulated in a voxel grid, ensuring spatial-temporal consistency. The universal voxel representation enables fusion with our \gls{cmst}.} 
    \label{3dlabels}
\end{figure}

Although point clouds are used for 2D-3D mapping, the semantic labels in our method originate from images. 
We hypothesize that 3D networks could provide complimentary information to our labels and thereby enhance label quality.
However, as pre-trained models are often unavailable for novel domains, we train a 3D segmentation module on the original camera-only pseudo-labels which we call a \glsreset{cmst} \gls{cmst}, illustrated in Fig.~\ref{3dlabels}$\textbf{C}$. 

We observe that voxels with a higher probability are generally more accurate, and hence hypothesize that we can use this to select a reliable set of pseudo-ground-truth labels for training. We select reliable labels by projecting scans onto the voxel grid and choosing a fixed percentage of the most confident (highest probability) labels per class. Although this reduces the number of supervision labels, \cite{genova_learning_2021, reichardt_360_2023} show that 3D networks can be effectively trained with limited labels. This approach ensures the model learns from the most reliable labels while maintaining class diversity. 
These reliable labels are used to fine-tune an arbitrary 3D point-cloud semantic segmentation network to provide domain-specific, 3D aware semantic input to our pseudo-labeling framework. These new predictions are then combined into the existing voxel grid using a Bayesian update. A temperature hyper-parameter is used to weigh the reliability of each input~\cite{hinton_distilling_2015}. 

Our \gls{cmst} scheme differs from traditional self-training in the 3D setting \cite{you_learning_2022, dao_label-efficient_2024} as it does not iteratively train a model using its own predictions as pseudo-labels. Rather, it trains a 3D perception model on the original camera-only labels, which is then used to improve those labels. 

\section{Experiments}

\renewcommand{\arraystretch}{1.3}
\setlength{\tabcolsep}{0.25em}
\begin{table*}[t]
\caption{Quality of our pseudo-labels compared to the ground-truth \textit{\textbf{val}} set on SemanticKITTI~\cite{behley_semantickitti_2019} and AgriUAV. \textbf{Best} and \underline{second best} results are marked. \ding{109} = point-wise 3D labels, \ding{111} = voxel-wise 3D labels. 3D-CN = 3D Consistency Network.}
\centering
\resizebox{\textwidth}{!}{%
\begin{tabular}{lccc||cc|ccccccccccc||c|ccccccc}
\hline \vspace{0.25em}
  & \multirow{2}{*}{\begin{turn}{90} 3D representation~ \end{turn}}  & \multirow{2}{*}{\begin{turn}{90}Bayesian fusion~~~ \end{turn}}  & \multirow{2}{*}{\begin{turn}{90}3D-CN iterations~~\end{turn}}  & \multicolumn{13}{c||}{\textbf{Automotive (SemanticKITTI\cite{behley_semantickitti_2019})}} & \multicolumn{8}{c}{\textbf{Aerial Vehicle (AgriUAV)}} \\ 
\textbf{Method}  &  &    &   & \begin{turn}{90}mIoU \%\end{turn} & \begin{turn}{90}cat. mIoU \%\end{turn} & \begin{turn}{90} \textcolor{car}{\ding{110}} car\end{turn} & \begin{turn}{90} \textcolor{bicycle}{\ding{110}} bicycle\end{turn}  & \begin{turn}{90} \textcolor{motorcycle}{\ding{110}} motorcycle\end{turn}    & \begin{turn}{90} \textcolor{other-vehicle}{\ding{110}} oth.-veh.\end{turn}  & \begin{turn}{90}\textcolor{person}{\ding{110}} person\end{turn}   & \begin{turn}{90}\textcolor{road}{\ding{110}} road\end{turn} & \begin{turn}{90}\textcolor{sidewalk}{\ding{110}} sidewalk\end{turn} & \begin{turn}{90}\textcolor{other-ground}{\ding{110}} oth.-ground\end{turn} & \begin{turn}{90}\textcolor{building}{\ding{110}} manmade\end{turn}  & \begin{turn}{90}\textcolor{vegetation}{\ding{110}} vegetation\end{turn} & \begin{turn}{90}\textcolor{terrain}{\ding{110}} terrain\end{turn}  & \begin{turn}{90}mIoU \%\end{turn} & \begin{turn}{90}\textcolor{vegetation}{\ding{110}} tree\end{turn} & \begin{turn}{90}\textcolor{pole}{\ding{110}} pole\end{turn}  & \begin{turn}{90}\textcolor{fence}{\ding{110}} fence\end{turn}  & \begin{turn}{90}\textcolor{wire}{\ding{110}} wire\end{turn}   & \begin{turn}{90}\textcolor{person}{\ding{110}} person\end{turn}  & \begin{turn}{90}\textcolor{building}{\ding{110}} building\end{turn}  & \begin{turn}{90}\textcolor{ground}{\ding{110}} ground\end{turn}
\\ \hline 
Pre-trained \cite{puy_using_2023} & \ding{109} & \ding{53} & - & 27.8 & 54.5 & 57.0 &	0.0 &	0.9 &	21.6 &	0.0 &	69.6 &	31.7 &	0.0 &	37.2 &	45.6 &	42.4 & 12.9 & 30.6 & 2.1 & 0.8 & - & 0.0 &  1.6 & 42.0\\
Ours (point)  & \ding{109}  & \ding{53} & - & 46.8 & 68.6 	&		77.2 &	\underline{25.5} &	15.1 &	\textbf{30.3} &	31.9 &	87.1 &	46.1 &	0.0 &	64.3 &	64.7 &	72.3 & 38.0 & 40.7 & 41.7 & 16.6 & \underline{2.9} & 7.8 & 83.8 & 72.1 \\
\textbf{Ours (voxel)}  & \ding{111} & \ding{51} & - & 48.9 & 71.2 &  86.1 &	20.4 &	22.1 &	\textbf{30.3} &	43.2 &	85.1 &	51.4 &	0.0 &	61.7 &	70.4 &	66.7 & 49.7 & 50.0 & 62.3 & 22.6 & \textbf{7.2} & \underline{31.8} & 95.7 & 78.6 \\
\textbf{~~+ 3D-CN} & \ding{111} & \ding{51} & 1 & \underline{57.6} & \underline{81.0} & 91.8 &	\textbf{25.7} &	\textbf{26.8} &	\underline{28.4} &	\underline{69.6} &	\underline{93.6} &	\underline{73.2} &	0.0 &	\underline{69.1} &	\underline{80.9} &	\underline{74.4} & \textbf{61.5} & \underline{82.5} &	\textbf{80.9} &	\underline{33.7} &	2.7 &	\textbf{39.8} &	\underline{96.5} &	\underline{94.5}\\   
\textbf{~~+ 3D-CN} & \ding{111}  & \ding{51} & 2 & \textbf{58.1} & \textbf{81.6} & \textbf{92.5} &	24.5 &	\textbf{26.8} &	27.3 &	\textbf{71.7} &	\textbf{93.9} &	\textbf{73.7} &	0.0 &	\textbf{69.4} &	\textbf{82.9} &	\textbf{76.2} & \underline{60.9} & \textbf{88.1} &	\underline{79.9} &	\textbf{41.5} &	0.4 &	{22.9} &	\textbf{96.8} &	\textbf{96.8}  
\\  \hline 
\end{tabular}%
}
\vspace*{-\baselineskip}
\label{table:main_compare}
\end{table*}

Our method's primary advantage lies in its capacity to generate labels for any set of classes in arbitrary domains, thus supporting novel applications where labeled 3D datasets are absent, such as \glspl{uav} and construction. However, to evaluate the quality of the labels quantitatively, we make use of two datasets with ground-truth labels.
First we assess  the quality of the pseudo-labels in both domains. Next, we evaluate their effectiveness in aiding domain adaptation. Finally, we illustrate how multi-model fusion through \gls{cmst} can further improve the quality of the pseudo-labels.

\subsection{Datasets}
\noindent \textbf{Automotive}: 
For the automotive domain, we utilize the widely used SemanticKITTI dataset~\cite{behley_semantickitti_2019} containing both RGB and LiDAR data with per-point semantic ground-truth labels. 

\noindent \textbf{UAV}:
To further demonstrate the versatility of our method across different domains, we create our own synthetic dataset with ground-truth labels in AirSim~\cite{shah_airsim_2017} which we call \textit{AgriUAV}. It includes seven classes relevant to agricultural applications and exhibits viewpoint variations common to \glspl{uav}. It contains RGB and LiDAR data, using a less common fixed-FOV solid-state LiDAR based on the Blickfeld Cube. It can thereby help to evaluate how our method generalizes across application and sensor domains.

\subsection{Implementation details}
\noindent \textbf{2D labels}: To generate 2D labels, we utilize the pre-trained foundation models Grounding Dino~\cite{liu_grounding_2023} and Segment Anything (SAM)~\cite{kirillov_segment_2023}. Specifically, we employ the Swin-T model for Grounding Dino~\cite{liu_grounding_2023} and the ViT-L model for SAM~\cite{kirillov_segment_2023}. We use $Q = 900$ and $T = 256$ for the number of query regions and text tokens respectively, and set the region similarity threshold of Grounding Dino~\cite{liu_grounding_2023} to 0.25 for SemanticKITTI~\cite{behley_semantickitti_2019}. For our synthetic dataset, we adjust the region similarity threshold to 0.2, as higher values resulted in very few masks on the synthetic camera images. For prompts, we expand the classes from the respective datasets as described in Section \ref{2dlabels}.

\noindent \textbf{3D labels}: We set the voxel size to 0.2 $m$ and accumulate all scans in a sequence in a single sparse voxel-grid. The final voxel-grid is further smoothed using $k=9$ nearest neighbors of each voxel.
To obtain pseudo-labels for evaluation, each LiDAR scan is projected onto the voxel grid. The class label of each point is determined by the class with the maximum probability within the corresponding voxel.

\noindent \textbf{3D Consistency Network}: For the \gls{cmst}, we employ WaffleIron~\cite{puy_using_2023} as our 3D backbone for its ease of implementation and adjustability.
We train the backbone using $XYZ$ coordinates as input features, supervised by the 20\% most confident original camera-based labels.
We train for a single epoch on the reliable pseudo-labels and fuse the output with the original sparse voxel-grid.

\subsection{Baselines}
\noindent \textbf{Label quality}: To assess label quality, we compare the pseudo-labels to the ground-truth labels of the dataset in question. 
Related labeling tools by \cite{zhou_openannotate2_2024} and \cite{zhou_openannotate3d_2024} are only available in limited capacity, labeling a maximum of 10 point clouds. Hence, as an automatic labeling baseline, we use a pre-trained segmentation model from a \textit{different} domain to mimic a scenario where new data needs to be labeled (indicated by 'Pre-trained'). For SemanticKITTI~\cite{behley_semantickitti_2019}, we use WaffleIron~\cite{puy_using_2023} pre-trained on nuScenes \cite{caesar_nuscenes_2020} and for AgriUAV, we use WaffleIron~\cite{puy_using_2023} pre-trained on SemanticKITTI~\cite{behley_semantickitti_2019}. We also compare our method with (\textbf{Ours (voxel)}) and without (Ours (point)) voxelization, projecting our 2D labels to 3D points. For a fair comparison, we limit evaluation to points within the camera frame and ignore unlabeled points.

\noindent \textbf{Domain adaptation}: We evaluate the output of WaffleIron~\cite{puy_using_2023} trained with different sets of labels. The oracle model is trained on the manually labeled ground-truth labels, whereas the source only model is trained on labels from another domain. The latter is then adapted to target domain using our automatically generated pseudo-labels, denoted with \textbf{Ours} for the camera-only version and \textbf{Ours + 3D-CN} for the version with \gls{cmst}.

\subsection{Metrics}
For quantitative evaluation on both datasets, we use the class-wise intersection over union (IoU) and the corresponding mean (mIoU) as our main metric for evaluation. To fairly compare the output of models across domains, we rename and merge several classes to pair models trained on nuScenes~\cite{caesar_nuscenes_2020} (16 classes) with SemanticKITTI~\cite{behley_semantickitti_2019} (19 classes) and AgriUAV (7 classes).
Additionally, following KITTI-360~\cite{liao_kitti-360_2023}, we also report category mIoU, where the 19 classes from SemanticKITTI~\cite{behley_semantickitti_2019} are grouped into 6 more coarse categories. This follows the observation that class descriptions from SemanticKITTI can be ambiguous even to a human annotator (e.g. the difference between 'terrain' and 'vegetation') and that the courser categories are usually sufficient for most semantic tasks.

\subsection{Pseudo-label quality}
\begin{figure*}
    \centering
    \includegraphics[width=\textwidth]{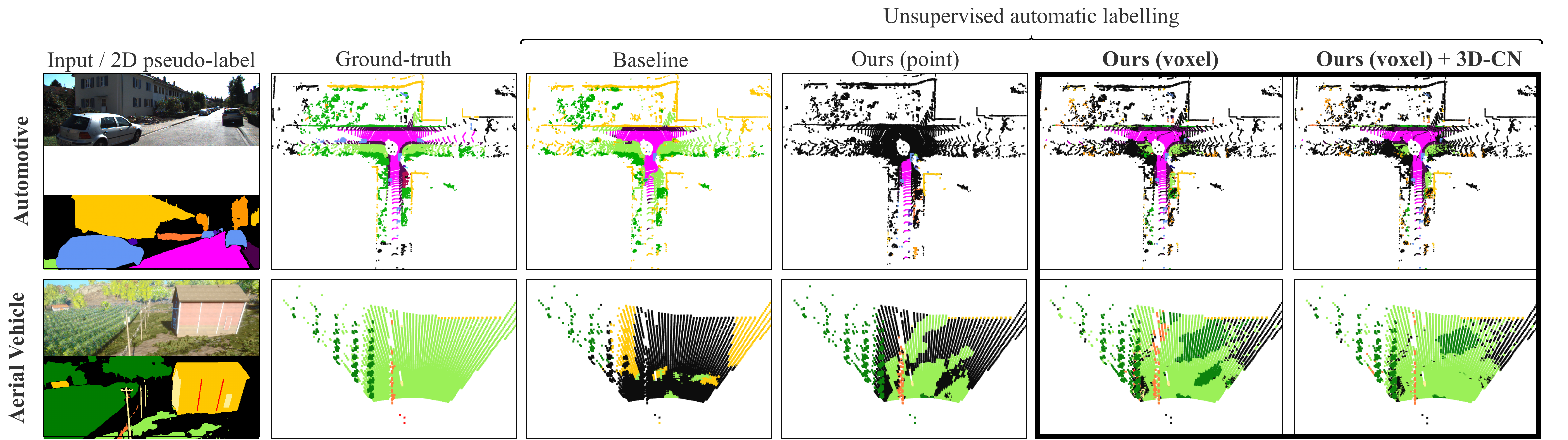}
    \caption{Qualitative results of our pseudo-labeling pipeline. Frames of SemanticKITTI~\cite{behley_semantickitti_2019} and our own AgriUAV drone dataset are shown on the top and bottom respectively. Colours correspond to the classes in Table \ref{table:domain_adapt}. Black points are unlabeled, e.g. when outside the camera frame or of an unsupported class. Voxels enable us to label vastly more points than 2D-3D projection alone. The bottom row clearly shows how pre-trained models from different domains often fail to transfer to new domains and how 3D-CN can improve spatial consistancy.} 
    \label{figure:qualitative}
\end{figure*}
We assess pseudo-label quality in various domains by comparing our automatically generated labels to the ground-truth labels on SemanticKITTI~\cite{behley_semantickitti_2019} and our AgriUAV drone dataset in Table~\ref{table:main_compare}. Ours (point) and \textbf{Ours (voxel)} outperform the pre-trained baseline considerably in both domains. It should be noted that the open-set ability of our method allows it to generate labels for all 19 classes of SemanticKITTI~\cite{behley_semantickitti_2019}, whereas the closed-set pre-trained model is limited to the classes from its source domain.
Whilst being vastly more memory efficient, our voxel-based (\textbf{Ours (voxel)}) method also outperforms the point-based (Ours (point)) version, despite a loss in resolution due to voxelization. By being able to efficiently accumulate and update the semantic voxels, we can label points that are unobserved by the camera. Hence, our voxel based approach labels over six times the number of points compared to the point-based version.
Fig.~\ref{figure:qualitative} shows this more clearly. 
Although our method improves label consistency,  we observe that the open-vocabulary \glspl{vfm} can struggle with ambiguous classes. 
For instance, it may split a moving bike into separate \textit{person} and \textit{bicycle} labels. Similar issues arise with highly ambiguous classes (e.g., \textit{road} versus \textit{other-ground}, \textit{terrain} versus \textit{vegetation}), which can be challenging even for expert human annotators. As a result, the coarser category mIoU shows a significant improvement.

\subsection{Domain adaptation to similar and new domains}
\label{section:domain_adapt}

\renewcommand{\arraystretch}{1.3}
\setlength{\tabcolsep}{0.25em}
\begin{table*}[t]
\caption{Inference results of the WaffleIron~\cite{puy_using_2023} segmentation backbone on the \textit{\textbf{val}} set of SemanticKITTI~\cite{behley_semantickitti_2019} and AgriUAV. The oracle model is trained on the ground-truth labels, whereas the source only model is trained on labels from another domain.}
\centering
\resizebox{0.95\textwidth}{!}{%
\begin{tabular}{clccc||c|ccccccccccc||c|ccccccc}
\hline \vspace{0.25em} 
  & & \multirow{2}{*}{\begin{turn}{90} Annotation free~~~ \end{turn}} & \multirow{2}{*}{\begin{turn}{90} Domain agnostic~~ \end{turn}} & \multirow{2}{*}{\begin{turn}{90} Any class~~~~~~~~~~ \end{turn}} & \multicolumn{12}{c||}{\textbf{Automotive (SemanticKITTI \cite{behley_semantickitti_2019})}} & \multicolumn{8}{c}{\textbf{Aerial Vehicle (AgriUAV)}} \\ 
& \textbf{Labels} & & & & \begin{turn}{90}mIoU \%\end{turn} & \begin{turn}{90} \textcolor{car}{\ding{110}} car\end{turn} & \begin{turn}{90} \textcolor{bicycle}{\ding{110}} bicycle\end{turn}  & \begin{turn}{90} \textcolor{motorcycle}{\ding{110}} motorcycle\end{turn}    & \begin{turn}{90} \textcolor{other-vehicle}{\ding{110}} oth.-veh.\end{turn}  & \begin{turn}{90}\textcolor{person}{\ding{110}} person\end{turn}   & \begin{turn}{90}\textcolor{road}{\ding{110}} road\end{turn} & \begin{turn}{90}\textcolor{sidewalk}{\ding{110}} sidewalk\end{turn} & \begin{turn}{90}\textcolor{other-ground}{\ding{110}} oth.-ground\end{turn} & \begin{turn}{90}\textcolor{building}{\ding{110}} manmade\end{turn}  & \begin{turn}{90}\textcolor{vegetation}{\ding{110}} vegetation\end{turn} & \begin{turn}{90}\textcolor{terrain}{\ding{110}} terrain\end{turn}  & \begin{turn}{90}mIoU \%\end{turn} & \begin{turn}{90}\textcolor{vegetation}{\ding{110}} tree\end{turn} & \begin{turn}{90}\textcolor{pole}{\ding{110}} pole\end{turn}  & \begin{turn}{90}\textcolor{fence}{\ding{110}} fence\end{turn}  & \begin{turn}{90}\textcolor{wire}{\ding{110}} wire\end{turn}   & \begin{turn}{90}\textcolor{person}{\ding{110}} person\end{turn}  & \begin{turn}{90}\textcolor{building}{\ding{110}} building\end{turn}  & \begin{turn}{90}\textcolor{ground}{\ding{110}} ground\end{turn}
\\ \hline 
\multirow{4}{*}{\parbox{2em}{\centering \begin{turn}{90} Wf.Iron \cite{puy_using_2023} \end{turn}}} & Oracle & \ding{53} & \ding{53} & \ding{53} & 76.0 & 95.9 &	81.4 &	78.4 &	70.3 &	82.3 &	94.7 &	80.4 &	2.8 &	90.5 &	90.6 &	72.3 &  59.1 & 78.0 &	64.9 &	45.6 &	68.1 &	3.1 &	63.7 &	90.2 
\\ \cdashline{2-25} 
&Source Only &\ding{51} & \ding{53} & \ding{53} & 30.6 & 68.2 &	0.0 &	5.4 &	\textbf{25.3} &	0.0 &	63.7 &	28.3 &	0.0 &	45.3 &	48.3 &	52.0 & 12.9 & 30.6 & 2.1 & 0.8 & - & 0.0 & 1.6 & 42.0 \\ 
&\textbf{Ours (voxel)} &\ding{51} & \ding{51} & \ding{51} & {42.1} & 89.4 &	7.6 &	11.8 &	16.5 &	\textbf{0.2} &	78.7 &	48.3 &	0.0 &	66.1 &	76.0 &	\textbf{68.6} & {33.4} & {40.3} & {48.2} & {9.2} & {16.6} & \textbf{1.4} & {40.4} & {77.4}\\
&\textbf{~~+ 3D-CN} &\ding{51} & \ding{51} & \ding{51} &  \textbf{47.9} & \textbf{92.0} &	\textbf{12.3} &	\textbf{32.4} &	{17.2} &	0.1 &	\textbf{88.8} &	\textbf{71.1} &	0.0 &	\textbf{68.1} &	\textbf{78.4} &	66.7 & \textbf{47.1} & \textbf{71.4} & \textbf{63.0} & \textbf{26.0} & \textbf{20.1} & {0.9} & \textbf{62.8} & \textbf{85.6}
\\  \hline 
\end{tabular}%
}
\vspace*{-\baselineskip}
\label{table:domain_adapt}
\end{table*}

A common issue with 3D LiDAR models is significantly reduced performance when a model trained on one dataset is evaluated on another, even when the classes are similar~\cite{rochan_unsupervised_2022}. When applications are different, such as using a model trained on automotive data on an \gls{uav}, this problem is further exacerbated as the target classes, viewpoints and sensors might differ significantly. To show the universal applicability and quality of our labels, we show how they can help in domain adaptation, even across different domains.

For the automotive domain, we evaluate the WaffleIron~\cite{puy_using_2023} model trained on nuScenes~\cite{caesar_nuscenes_2020} on the SemanticKITTI~\cite{behley_semantickitti_2019} \textbf{\textit{val}} set. Then, we use our method to generate pseudo-labels for sequence 00 of the \textbf{\textit{train}} set of SemanticKITTI~\cite{behley_semantickitti_2019} and fine-tune the model for a single epoch on those labels. 
For the \gls{uav} domain, we evaluate the WaffleIron~\cite{puy_using_2023} model trained on SemanticKITTI~\cite{behley_semantickitti_2019} on the \textbf{\textit{val}} set of our AgriUAV drone dataset. Then, we use our method to generate pseudo-labels for that dataset. To overcome the larger domain gap, we train the model for a longer 20 epochs on the pseudo-labeled \textbf{\textit{train}} set.

Table~\ref{table:domain_adapt} show the results. Naively using the Source Only model (i.e. trained on another dataset) degrades performance significantly, especially for minority classes, due to changes in vehicle, sensor and environment domains. This is also clearly shown in Fig.~\ref{figure:qualitative}. By fine-tuning on only a small number of pseudo-labels for just a single epoch we improve the mIoU of the original model by $11.5$ for the automotive domain. For the \gls{uav} domain, the domain gap is much larger, hence retraining the model for more epochs on our generated pseudo-labels results in a larger $20.5$ mIoU improvement. This demonstrates the ability of our method to provide a bridge to very different domains, applications and sensor setups\compressParag.

\subsection{3D Consistency Network}
\label{sec:self-training-results}
Finally, we investigate how the addition of the \gls{cmst} can enhance pseudo-label quality. As detailed in Section~\ref{sec:self-training}, we train a 3D segmentation backbone on our most confident image-based pseudo-labels and fuse the output with the original camera-based labels.

The last rows of Table~\ref{table:main_compare} show the results of \gls{cmst}. The quality of the pseudo-labels is improved significantly for almost all classes. Additionally, the point-wise output of the \gls{cmst} expands labeling capabilities beyond points observed by the camera, labeling all points.
Surprisingly, we observe that mIoU is higher for the fused pseudo-labels compared to \textit{either} our original camera-only pseudo-labels or the self-trained 3D model. We hypothesize that both modalities provide complementary information which enhances the combined pseudo-labels.
Multiple rounds of multi-modal self-training only show slight improvements. Additionally, we observe that the IoU goes \textit{down} for classes where the original IoU was already low, highlighting that self-training can potentially exacerbate mistakes. This is most apparent for underrepresented classes that are hard to observe by camera, like \textit{person}.
Training a model with these higher quality labels generally improves the performance of the model as well. When adapting a pre-trained automotive model to the \gls{uav} domain using our labels with \gls{cmst}, mIoU increases by $34.2$ compared to the unadapted model, approaching the oracle model trained on the ground-truth labels (shown in the bottom row of Table~\ref{table:domain_adapt}).

\section{Conclusion}
This work presents \textbf{LeAP}, a pseudo-labeling approach for 3D semantic tasks. By leveraging open-vocabulary foundation models, LeAP automatically generates consistent semantic 3D labels for any set of classes in any domain using only unlabeled image-pointcloud pairs as input. We propose a voxel based method that enable us to combine labels consistently over time through Bayesian updating, providing advantages in both label quality and quantity compared to other automatic labeling methods. We also introduce a 3D Consistency Network and show that it significantly enhances pseudo-label quality. Our method demonstrates versatility across various domains, tasks, and sensor configurations. The generated labels can be used to overcome domain gaps within and across diverse domains, with models trained for novel domains on our labels showing up to a $3.7\times$ improvement in mIoU compared to un-adapted baselines. Consequently, LeAP can help accelerate and expand the scope of 3D perception research into areas lacking labeled datasets, providing high-quality labels across various domains, diversifying the research field.

\noindent \textbf{Limitations and future work:} Although the use of 2D foundation models enables multi-domain labeling, we find that their output can be unpredictable, especially for ambiguous and highly specific classes. Hence, future work will focus on more advanced prompt engineering. 
Furthermore, as the \gls{cmst} is dependent on the quality of the original labels, it is prone to reinforce mistakes present in the original pseudo-labels. Self-training also cannot add new semantic information, so it is unable to correct systematic errors. To resolve this, future work will explore more advanced label selection mechanisms.
Finally, we currently do not differentiate between static and moving objects which can leave `tracks' in the voxel-grid. Although this rarely results in wrong labels, incorporating dynamics can further enhance label quality\compressParag.

{\small
\bibliographystyle{IEEEtran}
\bibliography{IEEEabrv,references}

\begin{thebibliography}{10}
\providecommand{\url}[1]{#1}
\csname url@rmstyle\endcsname
\providecommand{\newblock}{\relax}
\providecommand{\bibinfo}[2]{#2}
\providecommand\BIBentrySTDinterwordspacing{\spaceskip=0pt\relax}
\providecommand\BIBentryALTinterwordstretchfactor{4}
\providecommand\BIBentryALTinterwordspacing{\spaceskip=\fontdimen2\font plus
\BIBentryALTinterwordstretchfactor\fontdimen3\font minus \fontdimen4\font\relax}
\providecommand\BIBforeignlanguage[2]{{%
\expandafter\ifx\csname l@#1\endcsname\relax
\typeout{** WARNING: IEEEtran.bst: No hyphenation pattern has been}%
\typeout{** loaded for the language `#1'. Using the pattern for}%
\typeout{** the default language instead.}%
\else
\language=\csname l@#1\endcsname
\fi
#2}}

\bibitem{caesar_nuscenes_2020}
\BIBentryALTinterwordspacing
H.~Caesar, V.~Bankiti, A.~H. Lang, S.~Vora, V.~E. Liong, Q.~Xu, A.~Krishnan, Y.~Pan, G.~Baldan, and O.~Beijbom, ``\BIBforeignlanguage{en}{{nuScenes}: {A} {Multimodal} {Dataset} for {Autonomous} {Driving}},'' in \emph{\BIBforeignlanguage{en}{2020 {IEEE}/{CVF} {Conference} on {Computer} {Vision} and {Pattern} {Recognition} ({CVPR})}}.\hskip 1em plus 0.5em minus 0.4em\relax Seattle, WA, USA: IEEE, June 2020, pp. 11\,618--11\,628. [Online]. Available: \url{https://ieeexplore.ieee.org/document/9156412/}
\BIBentrySTDinterwordspacing

\bibitem{sun_scalability_2020}
\BIBentryALTinterwordspacing
P.~Sun, H.~Kretzschmar, X.~Dotiwalla, A.~Chouard, V.~Patnaik, P.~Tsui, J.~Guo, Y.~Zhou, Y.~Chai, B.~Caine, V.~Vasudevan, W.~Han, J.~Ngiam, H.~Zhao, A.~Timofeev, S.~Ettinger, M.~Krivokon, A.~Gao, A.~Joshi, Y.~Zhang, J.~Shlens, Z.~Chen, and D.~Anguelov, ``\BIBforeignlanguage{en}{Scalability in {Perception} for {Autonomous} {Driving}: {Waymo} {Open} {Dataset}},'' in \emph{\BIBforeignlanguage{en}{2020 {IEEE}/{CVF} {Conference} on {Computer} {Vision} and {Pattern} {Recognition} ({CVPR})}}.\hskip 1em plus 0.5em minus 0.4em\relax Seattle, WA, USA: IEEE, June 2020, pp. 2443--2451, arXiv:1912.04838 [cs, stat]. [Online]. Available: \url{https://ieeexplore.ieee.org/document/9156973/}
\BIBentrySTDinterwordspacing

\bibitem{geiger_are_2012}
\BIBentryALTinterwordspacing
A.~Geiger, P.~Lenz, and R.~Urtasun, ``\BIBforeignlanguage{en}{Are we ready for autonomous driving? {The} {KITTI} vision benchmark suite},'' in \emph{\BIBforeignlanguage{en}{2012 {IEEE} {Conference} on {Computer} {Vision} and {Pattern} {Recognition}}}.\hskip 1em plus 0.5em minus 0.4em\relax Providence, RI: IEEE, June 2012, pp. 3354--3361. [Online]. Available: \url{http://ieeexplore.ieee.org/document/6248074/}
\BIBentrySTDinterwordspacing

\bibitem{behley_semantickitti_2019}
\BIBentryALTinterwordspacing
J.~Behley, M.~Garbade, A.~Milioto, J.~Quenzel, S.~Behnke, C.~Stachniss, and J.~Gall, ``\BIBforeignlanguage{en}{{SemanticKITTI}: {A} {Dataset} for {Semantic} {Scene} {Understanding} of {LiDAR} {Sequences}},'' in \emph{\BIBforeignlanguage{en}{2019 {IEEE}/{CVF} {International} {Conference} on {Computer} {Vision} ({ICCV})}}.\hskip 1em plus 0.5em minus 0.4em\relax Seoul, Korea (South): IEEE, Oct. 2019, pp. 9296--9306, arXiv:1904.01416 [cs]. [Online]. Available: \url{https://ieeexplore.ieee.org/document/9010727/}
\BIBentrySTDinterwordspacing

\bibitem{liao_kitti-360_2023}
\BIBentryALTinterwordspacing
Y.~Liao, J.~Xie, and A.~Geiger, ``\BIBforeignlanguage{en}{{KITTI}-360: {A} {Novel} {Dataset} and {Benchmarks} for {Urban} {Scene} {Understanding} in {2D} and {3D}},'' \emph{\BIBforeignlanguage{en}{IEEE Transactions on Pattern Analysis and Machine Intelligence}}, vol.~45, no.~3, pp. 3292--3310, Mar. 2023, arXiv:2109.13410 [cs]. [Online]. Available: \url{https://ieeexplore.ieee.org/document/9786676/}
\BIBentrySTDinterwordspacing

\bibitem{zhu_ponderv2_2023}
\BIBentryALTinterwordspacing
H.~Zhu, H.~Yang, X.~Wu, D.~Huang, S.~Zhang, X.~He, T.~He, H.~Zhao, C.~Shen, Y.~Qiao, and W.~Ouyang, ``\BIBforeignlanguage{en}{{PonderV2}: {Pave} the {Way} for {3D} {Foundation} {Model} with {A} {Universal} {Pre}-training {Paradigm}},'' Oct. 2023, arXiv:2310.08586 [cs]. [Online]. Available: \url{http://arxiv.org/abs/2310.08586}
\BIBentrySTDinterwordspacing

\bibitem{liu_offline_2024}
\BIBentryALTinterwordspacing
X.~Liu and H.~Caesar, ``\BIBforeignlanguage{en}{Offline {Tracking} with {Object} {Permanence}},'' May 2024, arXiv:2310.01288 [cs]. [Online]. Available: \url{http://arxiv.org/abs/2310.01288}
\BIBentrySTDinterwordspacing

\bibitem{karnchanachari_towards_2024}
\BIBentryALTinterwordspacing
N.~Karnchanachari, D.~Geromichalos, K.~S. Tan, N.~Li, C.~Eriksen, S.~Yaghoubi, N.~Mehdipour, G.~Bernasconi, W.~K. Fong, Y.~Guo, and H.~Caesar, ``\BIBforeignlanguage{en}{Towards learning-based planning:{The} {nuPlan} benchmark for real-world autonomous driving},'' Mar. 2024, arXiv:2403.04133 [cs]. [Online]. Available: \url{http://arxiv.org/abs/2403.04133}
\BIBentrySTDinterwordspacing

\bibitem{radford_learning_2021}
\BIBentryALTinterwordspacing
A.~Radford, J.~W. Kim, C.~Hallacy, A.~Ramesh, G.~Goh, S.~Agarwal, G.~Sastry, A.~Askell, P.~Mishkin, J.~Clark, G.~Krueger, and I.~Sutskever, ``\BIBforeignlanguage{en}{Learning {Transferable} {Visual} {Models} {From} {Natural} {Language} {Supervision}},'' Feb. 2021, arXiv:2103.00020 [cs]. [Online]. Available: \url{http://arxiv.org/abs/2103.00020}
\BIBentrySTDinterwordspacing

\bibitem{kirillov_segment_2023}
\BIBentryALTinterwordspacing
A.~Kirillov, E.~Mintun, N.~Ravi, H.~Mao, C.~Rolland, L.~Gustafson, T.~Xiao, S.~Whitehead, A.~C. Berg, W.-Y. Lo, P.~Dollár, and R.~Girshick, ``\BIBforeignlanguage{en}{Segment {Anything}},'' in \emph{\BIBforeignlanguage{en}{2023 {IEEE}/{CVF} {International} {Conference} on {Computer} {Vision} ({ICCV})}}.\hskip 1em plus 0.5em minus 0.4em\relax Paris, France: IEEE, Oct. 2023, pp. 3992--4003, arXiv:2304.02643 [cs]. [Online]. Available: \url{https://ieeexplore.ieee.org/document/10378323/}
\BIBentrySTDinterwordspacing

\bibitem{yang_depth_2024}
\BIBentryALTinterwordspacing
L.~Yang, B.~Kang, Z.~Huang, X.~Xu, J.~Feng, and H.~Zhao, ``\BIBforeignlanguage{en}{Depth {Anything}: {Unleashing} the {Power} of {Large}-{Scale} {Unlabeled} {Data}},'' Jan. 2024, arXiv:2401.10891 [cs]. [Online]. Available: \url{http://arxiv.org/abs/2401.10891}
\BIBentrySTDinterwordspacing

\bibitem{zhang_pointclip_2022}
\BIBentryALTinterwordspacing
R.~Zhang, Z.~Guo, W.~Zhang, K.~Li, X.~Miao, B.~Cui, Y.~Qiao, P.~Gao, and H.~Li, ``\BIBforeignlanguage{en}{{PointCLIP}: {Point} {Cloud} {Understanding} by {CLIP}},'' in \emph{\BIBforeignlanguage{en}{2022 {IEEE}/{CVF} {Conference} on {Computer} {Vision} and {Pattern} {Recognition} ({CVPR})}}.\hskip 1em plus 0.5em minus 0.4em\relax New Orleans, LA, USA: IEEE, June 2022, pp. 8542--8552. [Online]. Available: \url{https://ieeexplore.ieee.org/document/9878980/}
\BIBentrySTDinterwordspacing

\bibitem{peng_openscene_2023}
\BIBentryALTinterwordspacing
S.~Peng, K.~Genova, C.~Jiang, A.~Tagliasacchi, M.~Pollefeys, and T.~Funkhouser, ``\BIBforeignlanguage{en}{{OpenScene}: {3D} {Scene} {Understanding} with {Open} {Vocabularies}},'' in \emph{\BIBforeignlanguage{en}{2023 {IEEE}/{CVF} {Conference} on {Computer} {Vision} and {Pattern} {Recognition} ({CVPR})}}.\hskip 1em plus 0.5em minus 0.4em\relax Vancouver, BC, Canada: IEEE, June 2023, pp. 815--824, arXiv:2211.15654 [cs]. [Online]. Available: \url{https://ieeexplore.ieee.org/document/10203983/}
\BIBentrySTDinterwordspacing

\bibitem{chen_clip2scene_2023}
\BIBentryALTinterwordspacing
R.~Chen, Y.~Liu, L.~Kong, X.~Zhu, Y.~Ma, Y.~Li, Y.~Hou, Y.~Qiao, and W.~Wang, ``\BIBforeignlanguage{en}{{CLIP2Scene}: {Towards} {Label}-efficient {3D} {Scene} {Understanding} by {CLIP}},'' in \emph{\BIBforeignlanguage{en}{2023 {IEEE}/{CVF} {Conference} on {Computer} {Vision} and {Pattern} {Recognition} ({CVPR})}}.\hskip 1em plus 0.5em minus 0.4em\relax Vancouver, BC, Canada: IEEE, June 2023, pp. 7020--7030, arXiv:2301.04926 [cs]. [Online]. Available: \url{https://ieeexplore.ieee.org/document/10204547/}
\BIBentrySTDinterwordspacing

\bibitem{tan_ovo_2023}
\BIBentryALTinterwordspacing
Z.~Tan, Z.~Dong, C.~Zhang, W.~Zhang, H.~Ji, and H.~Li, ``\BIBforeignlanguage{en}{{OVO}: {Open}-{Vocabulary} {Occupancy}},'' June 2023, arXiv:2305.16133 [cs]. [Online]. Available: \url{http://arxiv.org/abs/2305.16133}
\BIBentrySTDinterwordspacing

\bibitem{vobecky_pop-3d_2024}
\BIBentryALTinterwordspacing
A.~Vobecky, O.~Siméoni, D.~Hurych, S.~Gidaris, A.~Bursuc, P.~Pérez, and J.~Sivic, ``\BIBforeignlanguage{en}{{POP}-{3D}: {Open}-{Vocabulary} {3D} {Occupancy} {Prediction} from {Images}},'' Jan. 2024, arXiv:2401.09413 [cs]. [Online]. Available: \url{http://arxiv.org/abs/2401.09413}
\BIBentrySTDinterwordspacing

\bibitem{hess_lidarclip_2024}
\BIBentryALTinterwordspacing
G.~Hess, A.~Tonderski, C.~Petersson, K.~Åström, and L.~Svensson, ``\BIBforeignlanguage{en}{{LidarCLIP} or: {How} {I} {Learned} to {Talk} to {Point} {Clouds}},'' in \emph{\BIBforeignlanguage{en}{2024 {IEEE}/{CVF} {Winter} {Conference} on {Applications} of {Computer} {Vision} ({WACV})}}.\hskip 1em plus 0.5em minus 0.4em\relax Waikoloa, HI, USA: IEEE, Jan. 2024, pp. 7423--7432. [Online]. Available: \url{https://ieeexplore.ieee.org/document/10484207/}
\BIBentrySTDinterwordspacing

\bibitem{wu_towards_2023}
\BIBentryALTinterwordspacing
X.~Wu, Z.~Tian, X.~Wen, B.~Peng, X.~Liu, K.~Yu, and H.~Zhao, ``\BIBforeignlanguage{en}{Towards {Large}-scale {3D} {Representation} {Learning} with {Multi}-dataset {Point} {Prompt} {Training}},'' Aug. 2023, arXiv:2308.09718 [cs]. [Online]. Available: \url{http://arxiv.org/abs/2308.09718}
\BIBentrySTDinterwordspacing

\bibitem{liu_grounding_2023}
\BIBentryALTinterwordspacing
S.~Liu, Z.~Zeng, T.~Ren, F.~Li, H.~Zhang, J.~Yang, C.~Li, J.~Yang, H.~Su, J.~Zhu, and L.~Zhang, ``\BIBforeignlanguage{en}{Grounding {DINO}: {Marrying} {DINO} with {Grounded} {Pre}-{Training} for {Open}-{Set} {Object} {Detection}},'' Mar. 2023, arXiv:2303.05499 [cs]. [Online]. Available: \url{http://arxiv.org/abs/2303.05499}
\BIBentrySTDinterwordspacing

\bibitem{maturana_voxnet_2015}
\BIBentryALTinterwordspacing
D.~Maturana and S.~Scherer, ``\BIBforeignlanguage{en}{{VoxNet}: {A} {3D} {Convolutional} {Neural} {Network} for real-time object recognition},'' in \emph{\BIBforeignlanguage{en}{2015 {IEEE}/{RSJ} {International} {Conference} on {Intelligent} {Robots} and {Systems} ({IROS})}}.\hskip 1em plus 0.5em minus 0.4em\relax Hamburg, Germany: IEEE, Sept. 2015, pp. 922--928. [Online]. Available: \url{http://ieeexplore.ieee.org/document/7353481/}
\BIBentrySTDinterwordspacing

\bibitem{graham_3d_2018}
\BIBentryALTinterwordspacing
B.~Graham, M.~Engelcke, and L.~V.~D. Maaten, ``\BIBforeignlanguage{en}{{3D} {Semantic} {Segmentation} with {Submanifold} {Sparse} {Convolutional} {Networks}},'' in \emph{\BIBforeignlanguage{en}{2018 {IEEE}/{CVF} {Conference} on {Computer} {Vision} and {Pattern} {Recognition}}}.\hskip 1em plus 0.5em minus 0.4em\relax Salt Lake City, UT, USA: IEEE, June 2018, pp. 9224--9232, arXiv:1711.10275 [cs]. [Online]. Available: \url{https://ieeexplore.ieee.org/document/8579059/}
\BIBentrySTDinterwordspacing

\bibitem{vedaldi_searching_2020}
\BIBentryALTinterwordspacing
H.~Tang, Z.~Liu, S.~Zhao, Y.~Lin, J.~Lin, H.~Wang, and S.~Han, ``\BIBforeignlanguage{en}{Searching {Efficient} {3D} {Architectures} with {Sparse} {Point}-{Voxel} {Convolution}},'' \emph{\BIBforeignlanguage{en}{Computer Vision – ECCV 2020}}, vol. 12373, pp. 685--702, 2020, arXiv:2007.16100 [cs]. [Online]. Available: \url{https://link.springer.com/10.1007/978-3-030-58604-1_41}
\BIBentrySTDinterwordspacing

\bibitem{zhu_cylindrical_2021}
\BIBentryALTinterwordspacing
X.~Zhu, H.~Zhou, T.~Wang, F.~Hong, Y.~Ma, W.~Li, H.~Li, and D.~Lin, ``\BIBforeignlanguage{en}{Cylindrical and {Asymmetrical} {3D} {Convolution} {Networks} for {LiDAR} {Segmentation}},'' in \emph{\BIBforeignlanguage{en}{2021 {IEEE}/{CVF} {Conference} on {Computer} {Vision} and {Pattern} {Recognition} ({CVPR})}}.\hskip 1em plus 0.5em minus 0.4em\relax Nashville, TN, USA: IEEE, June 2021, pp. 9934--9943, arXiv:2011.10033 [cs]. [Online]. Available: \url{https://ieeexplore.ieee.org/document/9578697/}
\BIBentrySTDinterwordspacing

\bibitem{zhao_svaseg_2022}
\BIBentryALTinterwordspacing
L.~Zhao, S.~Xu, L.~Liu, D.~Ming, and W.~Tao, ``\BIBforeignlanguage{en}{{SVASeg}: {Sparse} {Voxel}-{Based} {Attention} for {3D} {LiDAR} {Point} {Cloud} {Semantic} {Segmentation}},'' \emph{\BIBforeignlanguage{en}{Remote Sensing}}, vol.~14, no.~18, p. 4471, Sept. 2022. [Online]. Available: \url{https://www.mdpi.com/2072-4292/14/18/4471}
\BIBentrySTDinterwordspacing

\bibitem{lai_spherical_2023}
\BIBentryALTinterwordspacing
X.~Lai, Y.~Chen, F.~Lu, J.~Liu, and J.~Jia, ``\BIBforeignlanguage{en}{Spherical {Transformer} for {LiDAR}-{Based} {3D} {Recognition}},'' in \emph{\BIBforeignlanguage{en}{2023 {IEEE}/{CVF} {Conference} on {Computer} {Vision} and {Pattern} {Recognition} ({CVPR})}}.\hskip 1em plus 0.5em minus 0.4em\relax Vancouver, BC, Canada: IEEE, June 2023, pp. 17\,545--17\,555, arXiv:2303.12766 [cs]. [Online]. Available: \url{https://ieeexplore.ieee.org/document/10203552/}
\BIBentrySTDinterwordspacing

\bibitem{zhang_occformer_2023}
\BIBentryALTinterwordspacing
Y.~Zhang, Z.~Zhu, and D.~Du, ``\BIBforeignlanguage{en}{{OccFormer}: {Dual}-path {Transformer} for {Vision}-based {3D} {Semantic} {Occupancy} {Prediction}},'' in \emph{\BIBforeignlanguage{en}{2023 {IEEE}/{CVF} {International} {Conference} on {Computer} {Vision} ({ICCV})}}.\hskip 1em plus 0.5em minus 0.4em\relax Paris, France: IEEE, Oct. 2023, pp. 9399--9409, arXiv:2304.05316 [cs]. [Online]. Available: \url{https://ieeexplore.ieee.org/document/10376645/}
\BIBentrySTDinterwordspacing

\bibitem{cao_monoscene_2022}
\BIBentryALTinterwordspacing
A.-Q. Cao and R.~De~Charette, ``\BIBforeignlanguage{en}{{MonoScene}: {Monocular} {3D} {Semantic} {Scene} {Completion}},'' in \emph{\BIBforeignlanguage{en}{2022 {IEEE}/{CVF} {Conference} on {Computer} {Vision} and {Pattern} {Recognition} ({CVPR})}}.\hskip 1em plus 0.5em minus 0.4em\relax New Orleans, LA, USA: IEEE, June 2022, pp. 3981--3991, arXiv:2112.00726 [cs]. [Online]. Available: \url{https://ieeexplore.ieee.org/document/9880217/}
\BIBentrySTDinterwordspacing

\bibitem{wu_squeezesegv2_2018}
\BIBentryALTinterwordspacing
B.~Wu, X.~Zhou, S.~Zhao, X.~Yue, and K.~Keutzer, ``\BIBforeignlanguage{en}{{SqueezeSegV2}: {Improved} {Model} {Structure} and {Unsupervised} {Domain} {Adaptation} for {Road}-{Object} {Segmentation} from a {LiDAR} {Point} {Cloud}},'' Sept. 2018, arXiv:1809.08495 [cs]. [Online]. Available: \url{http://arxiv.org/abs/1809.08495}
\BIBentrySTDinterwordspacing

\bibitem{milioto_rangenet_2019}
\BIBentryALTinterwordspacing
A.~Milioto, I.~Vizzo, J.~Behley, and C.~Stachniss, ``\BIBforeignlanguage{en}{{RangeNet} ++: {Fast} and {Accurate} {LiDAR} {Semantic} {Segmentation}},'' in \emph{\BIBforeignlanguage{en}{2019 {IEEE}/{RSJ} {International} {Conference} on {Intelligent} {Robots} and {Systems} ({IROS})}}.\hskip 1em plus 0.5em minus 0.4em\relax Macau, China: IEEE, Nov. 2019, pp. 4213--4220. [Online]. Available: \url{https://ieeexplore.ieee.org/document/8967762/}
\BIBentrySTDinterwordspacing

\bibitem{cortinhal_salsanext_2020}
\BIBentryALTinterwordspacing
T.~Cortinhal, G.~Tzelepis, and E.~E. Aksoy, ``\BIBforeignlanguage{en}{{SalsaNext}: {Fast}, {Uncertainty}-aware {Semantic} {Segmentation} of {LiDAR} {Point} {Clouds} for {Autonomous} {Driving}},'' July 2020, arXiv:2003.03653 [cs]. [Online]. Available: \url{http://arxiv.org/abs/2003.03653}
\BIBentrySTDinterwordspacing

\bibitem{zhang_polarnet_2020}
\BIBentryALTinterwordspacing
Y.~Zhang, Z.~Zhou, P.~David, X.~Yue, Z.~Xi, B.~Gong, and H.~Foroosh, ``\BIBforeignlanguage{en}{{PolarNet}: {An} {Improved} {Grid} {Representation} for {Online} {LiDAR} {Point} {Clouds} {Semantic} {Segmentation}},'' in \emph{\BIBforeignlanguage{en}{2020 {IEEE}/{CVF} {Conference} on {Computer} {Vision} and {Pattern} {Recognition} ({CVPR})}}.\hskip 1em plus 0.5em minus 0.4em\relax Seattle, WA, USA: IEEE, June 2020, pp. 9598--9607, arXiv:2003.14032 [cs]. [Online]. Available: \url{https://ieeexplore.ieee.org/document/9156460/}
\BIBentrySTDinterwordspacing

\bibitem{kong_rethinking_2023}
\BIBentryALTinterwordspacing
L.~Kong, Y.~Liu, R.~Chen, Y.~Ma, X.~Zhu, Y.~Li, Y.~Hou, Y.~Qiao, and Z.~Liu, ``\BIBforeignlanguage{en}{Rethinking {Range} {View} {Representation} for {LiDAR} {Segmentation}},'' in \emph{\BIBforeignlanguage{en}{2023 {IEEE}/{CVF} {International} {Conference} on {Computer} {Vision} ({ICCV})}}.\hskip 1em plus 0.5em minus 0.4em\relax Paris, France: IEEE, Oct. 2023, pp. 228--240, arXiv:2303.05367 [cs]. [Online]. Available: \url{https://ieeexplore.ieee.org/document/10376983/}
\BIBentrySTDinterwordspacing

\bibitem{roldao_lmscnet_2020}
\BIBentryALTinterwordspacing
L.~Roldao, R.~De~Charette, and A.~Verroust-Blondet, ``\BIBforeignlanguage{en}{{LMSCNet}: {Lightweight} {Multiscale} {3D} {Semantic} {Completion}},'' in \emph{\BIBforeignlanguage{en}{2020 {International} {Conference} on {3D} {Vision} ({3DV})}}.\hskip 1em plus 0.5em minus 0.4em\relax Fukuoka, Japan: IEEE, Nov. 2020, pp. 111--119, arXiv:2008.10559 [cs]. [Online]. Available: \url{https://ieeexplore.ieee.org/document/9320442/}
\BIBentrySTDinterwordspacing

\bibitem{cheng_s3cnet_2020}
\BIBentryALTinterwordspacing
R.~Cheng, C.~Agia, Y.~Ren, X.~Li, and L.~Bingbing, ``\BIBforeignlanguage{en}{{S3CNet}: {A} {Sparse} {Semantic} {Scene} {Completion} {Network} for {LiDAR} {Point} {Clouds}},'' Dec. 2020, arXiv:2012.09242 [cs]. [Online]. Available: \url{http://arxiv.org/abs/2012.09242}
\BIBentrySTDinterwordspacing

\bibitem{zuo_pointocc_2023}
\BIBentryALTinterwordspacing
S.~Zuo, W.~Zheng, Y.~Huang, J.~Zhou, and J.~Lu, ``\BIBforeignlanguage{en}{{PointOcc}: {Cylindrical} {Tri}-{Perspective} {View} for {Point}-based {3D} {Semantic} {Occupancy} {Prediction}},'' Aug. 2023, arXiv:2308.16896 [cs]. [Online]. Available: \url{http://arxiv.org/abs/2308.16896}
\BIBentrySTDinterwordspacing

\bibitem{huang_tri-perspective_2023}
\BIBentryALTinterwordspacing
Y.~Huang, W.~Zheng, Y.~Zhang, J.~Zhou, and J.~Lu, ``\BIBforeignlanguage{en}{Tri-{Perspective} {View} for {Vision}-{Based} {3D} {Semantic} {Occupancy} {Prediction}},'' in \emph{\BIBforeignlanguage{en}{2023 {IEEE}/{CVF} {Conference} on {Computer} {Vision} and {Pattern} {Recognition} ({CVPR})}}.\hskip 1em plus 0.5em minus 0.4em\relax Vancouver, BC, Canada: IEEE, June 2023, pp. 9223--9232, arXiv:2302.07817 [cs]. [Online]. Available: \url{https://ieeexplore.ieee.org/document/10203437/}
\BIBentrySTDinterwordspacing

\bibitem{qi_pointnet_2017}
\BIBentryALTinterwordspacing
C.~R. Qi, L.~Yi, H.~Su, and L.~J. Guibas, ``\BIBforeignlanguage{en}{{PointNet}++: {Deep} {Hierarchical} {Feature} {Learning} on {Point} {Sets} in a {Metric} {Space}},'' June 2017, arXiv:1706.02413 [cs]. [Online]. Available: \url{http://arxiv.org/abs/1706.02413}
\BIBentrySTDinterwordspacing

\bibitem{charles_pointnet_2017}
\BIBentryALTinterwordspacing
R.~Q. Charles, H.~Su, M.~Kaichun, and L.~J. Guibas, ``\BIBforeignlanguage{en}{{PointNet}: {Deep} {Learning} on {Point} {Sets} for {3D} {Classification} and {Segmentation}},'' in \emph{\BIBforeignlanguage{en}{2017 {IEEE} {Conference} on {Computer} {Vision} and {Pattern} {Recognition} ({CVPR})}}.\hskip 1em plus 0.5em minus 0.4em\relax Honolulu, HI: IEEE, July 2017, pp. 77--85, arXiv:1612.00593 [cs]. [Online]. Available: \url{http://ieeexplore.ieee.org/document/8099499/}
\BIBentrySTDinterwordspacing

\bibitem{thomas_kpconv_2019}
\BIBentryALTinterwordspacing
H.~Thomas, C.~R. Qi, J.-E. Deschaud, B.~Marcotegui, F.~Goulette, and L.~J. Guibas, ``\BIBforeignlanguage{en}{{KPConv}: {Flexible} and {Deformable} {Convolution} for {Point} {Clouds}},'' Aug. 2019, arXiv:1904.08889 [cs]. [Online]. Available: \url{http://arxiv.org/abs/1904.08889}
\BIBentrySTDinterwordspacing

\bibitem{hu_randla-net_2020}
\BIBentryALTinterwordspacing
Q.~Hu, B.~Yang, L.~Xie, S.~Rosa, Y.~Guo, Z.~Wang, N.~Trigoni, and A.~Markham, ``\BIBforeignlanguage{en}{{RandLA}-{Net}: {Efficient} {Semantic} {Segmentation} of {Large}-{Scale} {Point} {Clouds}},'' in \emph{\BIBforeignlanguage{en}{2020 {IEEE}/{CVF} {Conference} on {Computer} {Vision} and {Pattern} {Recognition} ({CVPR})}}.\hskip 1em plus 0.5em minus 0.4em\relax Seattle, WA, USA: IEEE, June 2020, pp. 11\,105--11\,114, arXiv:1911.11236 [cs, eess]. [Online]. Available: \url{https://ieeexplore.ieee.org/document/9156466/}
\BIBentrySTDinterwordspacing

\bibitem{wu_point_2022}
\BIBentryALTinterwordspacing
X.~Wu, Y.~Lao, L.~Jiang, X.~Liu, and H.~Zhao, ``\BIBforeignlanguage{en}{Point {Transformer} {V2}: {Grouped} {Vector} {Attention} and {Partition}-based {Pooling}},'' Oct. 2022, arXiv:2210.05666 [cs]. [Online]. Available: \url{http://arxiv.org/abs/2210.05666}
\BIBentrySTDinterwordspacing

\bibitem{cheng_af_2021}
\BIBentryALTinterwordspacing
R.~Cheng, R.~Razani, E.~Taghavi, E.~Li, and B.~Liu, ``\BIBforeignlanguage{en}{({AF}) $^{\textrm{2}}$ -{S3Net}: {Attentive} {Feature} {Fusion} with {Adaptive} {Feature} {Selection} for {Sparse} {Semantic} {Segmentation} {Network}},'' in \emph{\BIBforeignlanguage{en}{2021 {IEEE}/{CVF} {Conference} on {Computer} {Vision} and {Pattern} {Recognition} ({CVPR})}}.\hskip 1em plus 0.5em minus 0.4em\relax Nashville, TN, USA: IEEE, June 2021, pp. 12\,542--12\,551, arXiv:2102.04530 [cs]. [Online]. Available: \url{https://ieeexplore.ieee.org/document/9578725/}
\BIBentrySTDinterwordspacing

\bibitem{xu_rpvnet_2021}
\BIBentryALTinterwordspacing
J.~Xu, R.~Zhang, J.~Dou, Y.~Zhu, J.~Sun, and S.~Pu, ``\BIBforeignlanguage{en}{{RPVNet}: {A} {Deep} and {Efficient} {Range}-{Point}-{Voxel} {Fusion} {Network} for {LiDAR} {Point} {Cloud} {Segmentation}},'' in \emph{\BIBforeignlanguage{en}{2021 {IEEE}/{CVF} {International} {Conference} on {Computer} {Vision} ({ICCV})}}.\hskip 1em plus 0.5em minus 0.4em\relax Montreal, QC, Canada: IEEE, Oct. 2021, pp. 16\,004--16\,013, arXiv:2103.12978 [cs]. [Online]. Available: \url{https://ieeexplore.ieee.org/document/9709941/}
\BIBentrySTDinterwordspacing

\bibitem{ye_lidarmultinet_2023}
\BIBentryALTinterwordspacing
D.~Ye, Z.~Zhou, W.~Chen, Y.~Xie, Y.~Wang, P.~Wang, and H.~Foroosh, ``\BIBforeignlanguage{en}{{LidarMultiNet}: {Towards} a {Unified} {Multi}-{Task} {Network} for {LiDAR} {Perception}},'' Mar. 2023, arXiv:2209.09385 [cs]. [Online]. Available: \url{http://arxiv.org/abs/2209.09385}
\BIBentrySTDinterwordspacing

\bibitem{lu_lidar-camera_2024}
\BIBentryALTinterwordspacing
Z.~Lu, B.~Cao, and Q.~Hu, ``\BIBforeignlanguage{en}{{LiDAR}-{Camera} {Continuous} {Fusion} in {Voxelized} {Grid} for {Semantic} {Scene} {Completion}},'' \emph{\BIBforeignlanguage{en}{IEEE Transactions on Circuits and Systems for Video Technology}}, pp. 1--1, 2024. [Online]. Available: \url{https://ieeexplore.ieee.org/document/10613892/}
\BIBentrySTDinterwordspacing

\bibitem{zhang_growsp_2023}
\BIBentryALTinterwordspacing
Z.~Zhang, B.~Yang, B.~Wang, and B.~Li, ``\BIBforeignlanguage{en}{{GrowSP}: {Unsupervised} {Semantic} {Segmentation} of {3D} {Point} {Clouds}},'' in \emph{\BIBforeignlanguage{en}{2023 {IEEE}/{CVF} {Conference} on {Computer} {Vision} and {Pattern} {Recognition} ({CVPR})}}.\hskip 1em plus 0.5em minus 0.4em\relax Vancouver, BC, Canada: IEEE, June 2023, pp. 17\,619--17\,629, arXiv:2305.16404 [cs]. [Online]. Available: \url{https://ieeexplore.ieee.org/document/10203698/}
\BIBentrySTDinterwordspacing

\bibitem{liu_u3ds3_nodate}
J.~Liu, Z.~Yu, T.~P. Breckon, and H.~P.~H. Shum, ``\BIBforeignlanguage{en}{{U3DS3}: {Unsupervised} {3D} {Semantic} {Scene} {Segmentation}},'' arXiv:2311.06018 [cs].

\bibitem{xie_pointcontrast_2020}
\BIBentryALTinterwordspacing
S.~Xie, J.~Gu, D.~Guo, C.~R. Qi, L.~J. Guibas, and O.~Litany, ``\BIBforeignlanguage{en}{{PointContrast}: {Unsupervised} {Pre}-training for {3D} {Point} {Cloud} {Understanding}},'' Nov. 2020, arXiv:2007.10985 [cs]. [Online]. Available: \url{http://arxiv.org/abs/2007.10985}
\BIBentrySTDinterwordspacing

\bibitem{zhang_self-supervised_2021}
\BIBentryALTinterwordspacing
Z.~Zhang, R.~Girdhar, A.~Joulin, and I.~Misra, ``\BIBforeignlanguage{en}{Self-{Supervised} {Pretraining} of {3D} {Features} on any {Point}-{Cloud}},'' in \emph{\BIBforeignlanguage{en}{2021 {IEEE}/{CVF} {International} {Conference} on {Computer} {Vision} ({ICCV})}}.\hskip 1em plus 0.5em minus 0.4em\relax Montreal, QC, Canada: IEEE, Oct. 2021, pp. 10\,232--10\,243. [Online]. Available: \url{https://ieeexplore.ieee.org/document/9710368/}
\BIBentrySTDinterwordspacing

\bibitem{nunes_segcontrast_2022}
\BIBentryALTinterwordspacing
L.~Nunes, R.~Marcuzzi, X.~Chen, J.~Behley, and C.~Stachniss, ``\BIBforeignlanguage{en}{{SegContrast}: {3D} {Point} {Cloud} {Feature} {Representation} {Learning} {Through} {Self}-{Supervised} {Segment} {Discrimination}},'' \emph{\BIBforeignlanguage{en}{IEEE Robotics and Automation Letters}}, vol.~7, no.~2, pp. 2116--2123, Apr. 2022. [Online]. Available: \url{https://ieeexplore.ieee.org/document/9681336/}
\BIBentrySTDinterwordspacing

\bibitem{hayler_s4c_2024}
\BIBentryALTinterwordspacing
A.~Hayler, F.~Wimbauer, D.~Muhle, C.~Rupprecht, and D.~Cremers, ``\BIBforeignlanguage{en}{{S4C}: {Self}-{Supervised} {Semantic} {Scene} {Completion} {With} {Neural} {Fields}},'' in \emph{\BIBforeignlanguage{en}{2024 {International} {Conference} on {3D} {Vision} ({3DV})}}.\hskip 1em plus 0.5em minus 0.4em\relax Davos, Switzerland: IEEE, Mar. 2024, pp. 409--420, arXiv:2310.07522 [cs]. [Online]. Available: \url{https://ieeexplore.ieee.org/document/10550759/}
\BIBentrySTDinterwordspacing

\bibitem{zhang_occnerf_2023}
\BIBentryALTinterwordspacing
C.~Zhang, J.~Yan, Y.~Wei, J.~Li, L.~Liu, Y.~Tang, Y.~Duan, and J.~Lu, ``\BIBforeignlanguage{en}{{OccNeRF}: {Self}-{Supervised} {Multi}-{Camera} {Occupancy} {Prediction} with {Neural} {Radiance} {Fields}},'' Dec. 2023, arXiv:2312.09243 [cs]. [Online]. Available: \url{http://arxiv.org/abs/2312.09243}
\BIBentrySTDinterwordspacing

\bibitem{genova_learning_2021}
\BIBentryALTinterwordspacing
K.~Genova, X.~Yin, A.~Kundu, C.~Pantofaru, F.~Cole, A.~Sud, B.~Brewington, B.~Shucker, and T.~Funkhouser, ``\BIBforeignlanguage{en}{Learning {3D} {Semantic} {Segmentation} with only {2D} {Image} {Supervision}},'' in \emph{\BIBforeignlanguage{en}{2021 {International} {Conference} on {3D} {Vision} ({3DV})}}.\hskip 1em plus 0.5em minus 0.4em\relax London, United Kingdom: IEEE, Dec. 2021, pp. 361--372. [Online]. Available: \url{https://ieeexplore.ieee.org/document/9665849/}
\BIBentrySTDinterwordspacing

\bibitem{bultmann_real-time_2023}
\BIBentryALTinterwordspacing
S.~Bultmann, J.~Quenzel, and S.~Behnke, ``\BIBforeignlanguage{en}{Real-time multi-modal semantic fusion on unmanned aerial vehicles with label propagation for cross-domain adaptation},'' \emph{\BIBforeignlanguage{en}{Robotics and Autonomous Systems}}, vol. 159, p. 104286, Jan. 2023. [Online]. Available: \url{https://linkinghub.elsevier.com/retrieve/pii/S0921889022001750}
\BIBentrySTDinterwordspacing

\bibitem{sautier_image--lidar_2022}
\BIBentryALTinterwordspacing
C.~Sautier, G.~Puy, S.~Gidaris, A.~Boulch, A.~Bursuc, and R.~Marlet, ``\BIBforeignlanguage{en}{Image-to-{Lidar} {Self}-{Supervised} {Distillation} for {Autonomous} {Driving} {Data}},'' in \emph{\BIBforeignlanguage{en}{2022 {IEEE}/{CVF} {Conference} on {Computer} {Vision} and {Pattern} {Recognition} ({CVPR})}}.\hskip 1em plus 0.5em minus 0.4em\relax New Orleans, LA, USA: IEEE, June 2022, pp. 9881--9891. [Online]. Available: \url{https://ieeexplore.ieee.org/document/9879430/}
\BIBentrySTDinterwordspacing

\bibitem{mahmoud_self-supervised_2023}
\BIBentryALTinterwordspacing
A.~Mahmoud, J.~S.~K. Hu, T.~Kuai, A.~Harakeh, L.~Paull, and S.~L. Waslander, ``\BIBforeignlanguage{en}{Self-{Supervised} {Image}-to-{Point} {Distillation} via {Semantically} {Tolerant} {Contrastive} {Loss}},'' in \emph{\BIBforeignlanguage{en}{2023 {IEEE}/{CVF} {Conference} on {Computer} {Vision} and {Pattern} {Recognition} ({CVPR})}}.\hskip 1em plus 0.5em minus 0.4em\relax Vancouver, BC, Canada: IEEE, June 2023, pp. 7102--7110, arXiv:2301.05709 [cs]. [Online]. Available: \url{https://ieeexplore.ieee.org/document/10204499/}
\BIBentrySTDinterwordspacing

\bibitem{liu_segment_2023}
\BIBentryALTinterwordspacing
Y.~Liu, L.~Kong, J.~Cen, R.~Chen, W.~Zhang, L.~Pan, K.~Chen, and Z.~Liu, ``\BIBforeignlanguage{en}{Segment {Any} {Point} {Cloud} {Sequences} by {Distilling} {Vision} {Foundation} {Models}},'' Oct. 2023, arXiv:2306.09347 [cs]. [Online]. Available: \url{http://arxiv.org/abs/2306.09347}
\BIBentrySTDinterwordspacing

\bibitem{vora_pointpainting_2020}
\BIBentryALTinterwordspacing
S.~Vora, A.~H. Lang, B.~Helou, and O.~Beijbom, ``\BIBforeignlanguage{en}{{PointPainting}: {Sequential} {Fusion} for {3D} {Object} {Detection}},'' in \emph{\BIBforeignlanguage{en}{2020 {IEEE}/{CVF} {Conference} on {Computer} {Vision} and {Pattern} {Recognition} ({CVPR})}}.\hskip 1em plus 0.5em minus 0.4em\relax Seattle, WA, USA: IEEE, June 2020, pp. 4603--4611, arXiv:1911.10150 [cs, eess, stat]. [Online]. Available: \url{https://ieeexplore.ieee.org/document/9156790/}
\BIBentrySTDinterwordspacing

\bibitem{reichardt_360_2023}
\BIBentryALTinterwordspacing
L.~Reichardt, N.~Ebert, and O.~Wasenmüller, ``\BIBforeignlanguage{en}{360° from a {Single} {Camera}: {A} {Few}-{Shot} {Approach} for {LiDAR} {Segmentation}},'' in \emph{\BIBforeignlanguage{en}{2023 {IEEE}/{CVF} {International} {Conference} on {Computer} {Vision} {Workshops} ({ICCVW})}}.\hskip 1em plus 0.5em minus 0.4em\relax Paris, France: IEEE, Oct. 2023, pp. 1067--1075, arXiv:2309.06197 [cs]. [Online]. Available: \url{https://ieeexplore.ieee.org/document/10350853/}
\BIBentrySTDinterwordspacing

\bibitem{khurana_shelf-supervised_2024}
\BIBentryALTinterwordspacing
M.~Khurana, N.~Peri, D.~Ramanan, and J.~Hays, ``\BIBforeignlanguage{en}{Shelf-{Supervised} {Multi}-{Modal} {Pre}-{Training} for {3D} {Object} {Detection}},'' June 2024, arXiv:2406.10115 [cs]. [Online]. Available: \url{http://arxiv.org/abs/2406.10115}
\BIBentrySTDinterwordspacing

\bibitem{zhang_sam3d_2024}
\BIBentryALTinterwordspacing
D.~Zhang, D.~Liang, H.~Yang, Z.~Zou, X.~Ye, Z.~Liu, and X.~Bai, ``\BIBforeignlanguage{en}{{SAM3D}: zero-shot {3D} object detection via the segment anything model},'' \emph{\BIBforeignlanguage{en}{Science China Information Sciences}}, vol.~67, no.~4, p. 149101, Mar. 2024, arXiv:2306.02245 [cs, eess]. [Online]. Available: \url{http://arxiv.org/abs/2306.02245}
\BIBentrySTDinterwordspacing

\bibitem{najibi_unsupervised_2023}
\BIBentryALTinterwordspacing
M.~Najibi, J.~Ji, Y.~Zhou, C.~R. Qi, X.~Yan, S.~Ettinger, and D.~Anguelov, ``\BIBforeignlanguage{en}{Unsupervised {3D} {Perception} with {2D} {Vision}-{Language} {Distillation} for {Autonomous} {Driving}},'' in \emph{\BIBforeignlanguage{en}{2023 {IEEE}/{CVF} {International} {Conference} on {Computer} {Vision} ({ICCV})}}.\hskip 1em plus 0.5em minus 0.4em\relax Paris, France: IEEE, Oct. 2023, pp. 8568--8578, arXiv:2309.14491 [cs]. [Online]. Available: \url{https://ieeexplore.ieee.org/document/10377030/}
\BIBentrySTDinterwordspacing

\bibitem{zhou_openannotate3d_2024}
\BIBentryALTinterwordspacing
Y.~Zhou, L.~Cai, X.~Cheng, Z.~Gan, X.~Xue, and W.~Ding, ``\BIBforeignlanguage{en}{{OpenAnnotate3D}: {Open}-{Vocabulary} {Auto}-{Labeling} {System} for {Multi}-modal {3D} {Data}},'' in \emph{\BIBforeignlanguage{en}{2024 {IEEE} {International} {Conference} on {Robotics} and {Automation} ({ICRA})}}.\hskip 1em plus 0.5em minus 0.4em\relax Yokohama, Japan: IEEE, May 2024, pp. 9086--9092. [Online]. Available: \url{https://ieeexplore.ieee.org/document/10610779/}
\BIBentrySTDinterwordspacing

\bibitem{zhou_openannotate2_2024}
\BIBentryALTinterwordspacing
Y.~Zhou, L.~Cai, X.~Cheng, Q.~Zhang, X.~Xue, W.~Ding, and J.~Pu, ``\BIBforeignlanguage{en}{{OpenAnnotate2}: {Multi}-{Modal} {Auto}-{Annotating} for {Autonomous} {Driving}},'' \emph{\BIBforeignlanguage{en}{IEEE Transactions on Intelligent Vehicles}}, pp. 1--13, 2024. [Online]. Available: \url{https://ieeexplore.ieee.org/document/10480248/}
\BIBentrySTDinterwordspacing

\bibitem{mccormac_semanticfusion_2017}
\BIBentryALTinterwordspacing
J.~McCormac, A.~Handa, A.~Davison, and S.~Leutenegger, ``\BIBforeignlanguage{en}{{SemanticFusion}: {Dense} {3D} semantic mapping with convolutional neural networks},'' in \emph{\BIBforeignlanguage{en}{2017 {IEEE} {International} {Conference} on {Robotics} and {Automation} ({ICRA})}}.\hskip 1em plus 0.5em minus 0.4em\relax Singapore, Singapore: IEEE, May 2017, pp. 4628--4635, arXiv:1609.05130 [cs]. [Online]. Available: \url{http://ieeexplore.ieee.org/document/7989538/}
\BIBentrySTDinterwordspacing

\bibitem{niesner_real-time_2013}
\BIBentryALTinterwordspacing
M.~Nießner, M.~Zollhöfer, S.~Izadi, and M.~Stamminger, ``\BIBforeignlanguage{en}{Real-time {3D} reconstruction at scale using voxel hashing},'' \emph{\BIBforeignlanguage{en}{ACM Transactions on Graphics}}, vol.~32, no.~6, pp. 1--11, Nov. 2013. [Online]. Available: \url{https://dl.acm.org/doi/10.1145/2508363.2508374}
\BIBentrySTDinterwordspacing

\bibitem{hinton_distilling_2015}
\BIBentryALTinterwordspacing
G.~Hinton, O.~Vinyals, and J.~Dean, ``\BIBforeignlanguage{en}{Distilling the {Knowledge} in a {Neural} {Network}},'' Mar. 2015, arXiv:1503.02531 [cs, stat]. [Online]. Available: \url{http://arxiv.org/abs/1503.02531}
\BIBentrySTDinterwordspacing

\bibitem{you_learning_2022}
\BIBentryALTinterwordspacing
Y.~You, K.~Luo, C.~P. Phoo, W.-L. Chao, W.~Sun, B.~Hariharan, M.~Campbell, and K.~Q. Weinberger, ``\BIBforeignlanguage{en}{Learning to {Detect} {Mobile} {Objects} from {LiDAR} {Scans} {Without} {Labels}},'' in \emph{\BIBforeignlanguage{en}{2022 {IEEE}/{CVF} {Conference} on {Computer} {Vision} and {Pattern} {Recognition} ({CVPR})}}.\hskip 1em plus 0.5em minus 0.4em\relax New Orleans, LA, USA: IEEE, June 2022, pp. 1120--1130. [Online]. Available: \url{https://ieeexplore.ieee.org/document/9879816/}
\BIBentrySTDinterwordspacing

\bibitem{dao_label-efficient_2024}
\BIBentryALTinterwordspacing
M.-Q. Dao, H.~Caesar, J.~S. Berrio, M.~Shan, S.~Worrall, V.~Frémont, and E.~Malis, ``\BIBforeignlanguage{en}{Label-{Efficient} {3D} {Object} {Detection} {For} {Road}-{Side} {Units}},'' Apr. 2024, arXiv:2404.06256 [cs]. [Online]. Available: \url{http://arxiv.org/abs/2404.06256}
\BIBentrySTDinterwordspacing

\bibitem{puy_using_2023}
\BIBentryALTinterwordspacing
G.~Puy, A.~Boulch, and R.~Marlet, ``\BIBforeignlanguage{en}{Using a {Waffle} {Iron} for {Automotive} {Point} {Cloud} {Semantic} {Segmentation}},'' in \emph{\BIBforeignlanguage{en}{2023 {IEEE}/{CVF} {International} {Conference} on {Computer} {Vision} ({ICCV})}}.\hskip 1em plus 0.5em minus 0.4em\relax Paris, France: IEEE, Oct. 2023, pp. 3356--3366, arXiv:2301.10100 [cs]. [Online]. Available: \url{https://ieeexplore.ieee.org/document/10378314/}
\BIBentrySTDinterwordspacing

\bibitem{shah_airsim_2017}
\BIBentryALTinterwordspacing
S.~Shah, D.~Dey, C.~Lovett, and A.~Kapoor, ``\BIBforeignlanguage{en}{{AirSim}: {High}-{Fidelity} {Visual} and {Physical} {Simulation} for {Autonomous} {Vehicles}},'' July 2017, arXiv:1705.05065 [cs]. [Online]. Available: \url{http://arxiv.org/abs/1705.05065}
\BIBentrySTDinterwordspacing

\bibitem{rochan_unsupervised_2022}
\BIBentryALTinterwordspacing
M.~Rochan, S.~Aich, E.~R. Corral-Soto, A.~Nabatchian, and B.~Liu, ``\BIBforeignlanguage{en}{Unsupervised {Domain} {Adaptation} in {LiDAR} {Semantic} {Segmentation} with {Self}-{Supervision} and {Gated} {Adapters}},'' in \emph{\BIBforeignlanguage{en}{2022 {International} {Conference} on {Robotics} and {Automation} ({ICRA})}}.\hskip 1em plus 0.5em minus 0.4em\relax Philadelphia, PA, USA: IEEE, May 2022, pp. 2649--2655, arXiv:2107.09783 [cs]. [Online]. Available: \url{https://ieeexplore.ieee.org/document/9811654/}
\BIBentrySTDinterwordspacing

\end{thebibliography}
}

\end{document}